\pdfoutput=1
\documentclass{article}

\usepackage{microtype}
\usepackage{graphicx}
\usepackage{subfigure}
\usepackage{booktabs} % for professional tables

\usepackage[utf8]{inputenc} % allow utf-8 input
\usepackage[T1]{fontenc}    % use 8-bit T1 fonts
\usepackage{url}            % simple URL typesetting
\usepackage{booktabs}       % professional-quality tables
\usepackage{amsfonts}       % blackboard math symbols
\usepackage{nicefrac}       % compact symbols for 1/2, etc.
\usepackage{microtype}      % microtypography
\usepackage{xcolor}         % colors
\usepackage{adjustbox} % in preamble

\usepackage{amsmath}
\usepackage{graphicx}

\usepackage{hyperref}

\usepackage[accepted]{mlsys2025}

\mlsystitlerunning{SLM finetuning for NL to Code}

\begin{document}

\twocolumn[
\mlsystitle{SLM Finetuning For Natural Language To Domain Specific Code Generation In Production}

% List of affiliations: The first argument should be a (short)
% identifier you will use later to specify author affiliations
% Academic affiliations should list Department, University, City, Region, Country
% Industry affiliations should list Company, City, Region, Country

% You can specify symbols, otherwise they are numbered in order.
% Ideally, you should not use this facility. Affiliations will be numbered
% in order of appearance and this is the preferred way.
\mlsyssetsymbol{equal}{*}

\begin{mlsysauthorlist}
\mlsysauthor{Renjini R. Nair}{ms}
\mlsysauthor{Damian K. Kowalczyk}{ms}
\mlsysauthor{Marco Gaudesi}{ms}
\mlsysauthor{Chhaya Methani}{ms}
\end{mlsysauthorlist}

\mlsysaffiliation{ms}{Microsoft Research, Seattle, WA, USA}

\mlsyscorrespondingauthor{Renjini R. Nair}{reramada@microsoft.com}
\mlsyscorrespondingauthor{Chhaya Methani}{cmethani@microsoft.com}

% You may provide any keywords that you
% find helpful for describing your paper; these are used to populate
% the "keywords" metadata in the PDF but will not be shown in the document
\mlsyskeywords{SLM, Finetuning, Domain Specific Language}

\vskip 0.3in

\begin{abstract}
Many applications today use LLMs for code generation, however, production systems have strict latency requirements that can be hard to meet with large models. Small language models (SLMs), with a few billion parameters, are resource-efficient, but might suffer from limited reasoning, hallucinations or poor retention of longer context. They are 50-fold smaller than LLMs, but the smaller size can result in brittle outputs and a lower performance ceiling, having retained less knowledge and context. Fine-tuning improves accuracy for specific tasks by embedding domain knowledge directly into model weights, reducing reliance on runtime context. We had previously implemented a baseline NL-to-Code generation approach using a retrieval-augmented generation (RAG) pipeline that dynamically selected few-shot examples to embed DSL context for the LLM. In the current study, we evaluate SLMs for generating this DSL from natural language by fine-tuning variants of Mistral and other SLMs on a dataset of NL-code pairs, aiming to improve upon the baseline. Our results show that the fine-tuned models achieve better performance and latency for test data sets than the larger models. The best performing SLM could be directly augmented by finetuning with minimal amounts of various classes of harmful data, enabling it to be Responsible AI compliant. We also present an experiment where we further fine-tuned the trained SLM for customer-specific scenarios without general performance degradation, which would help to resolve customer specific issues that can arise in production systems. For a well-rounded study, we performed load testing, followed by deployment to production, confirming optimal performance in terms of latency and quality. Our findings show that task-specific finetuning with SLMs provides an efficient, faster, cost-effective and adaptive alternative to LLMs for domain specific language (DSL) generation.
\end{abstract}
]

% ================= PREPRINT + IEEE NOTICE =================
\begingroup
\renewcommand\thefootnote{}
\footnotetext{
\footnotesize
\textbf{Preprint.} A preprint version of this work is available at arXiv.
}
\endgroup
% =========================================================

\section{Introduction}

The landscape of natural language processing (NLP) is rapidly evolving with advances in large language models (LLMs) and the growing utility of small language models (SLMs). LLMs, with billions of parameters, have demonstrated strong generalization across tasks such as open-ended generation, multi-hop reasoning, and few-shot learning  \cite{brown2020language, chowdhery2022palm}. However, LLMs present considerable challenges in terms of computational cost, inference latency, model interpretability, and deployment constraints, especially in security and privacy sensitive or resource-constrained settings \cite{chen2025surveyprivacyrisksprotection, staab2024memorizationviolatingprivacyinference}. Small Language Models (SLMs), by contrast, offer a more lightweight and tractable alternative, particularly well-suited for task-specific or edge applications \cite{zhang2023opt}. While the reduced size of SLMs enables faster inference and easier deployment, it also limits their internal knowledge and reasoning depth, often resulting in brittle performance, increased hallucinations, and sensitivity to prompt formulation \cite{subramanian2025smalllanguagemodelsslms, halo2023hallucinations, wee2024hallucination, min2023calibrated}. In short, LLMs are powerful generalists, while SLMs are efficient specialists.

To mitigate accuracy limitations of SLMs, fine-tuning on domain-specific datasets has emerged as an effective strategy \cite{bi2024enhancing, min2022rethinking, yu2021differentially, bappy2025case}. By adapting model weights to reflect task-relevant structure and language, fine-tuning can substantially improve accuracy and output consistency \cite{xu2023small}. Parameter-efficient finetuning (PEFT) approaches such as Low-Rank Adaptation (LoRA) and QLoRA enable this adaptation without retraining the entire model, reducing both memory and storage overhead \cite{hu2021lora}. These methods introduce trainable low-rank matrices within existing transformer layers, making it feasible to deploy customized models on consumer-grade hardware. In this paper, we fine-tune SLMs using PEFT w/LoRA for a structured task with stable domain boundaries. Compared to retrieval-augmented generation (RAG) with LLMs, which requires maintaining external document stores and adds inference latency, finetuning allows to reduce the context necessary per request, and consequently, prefill-related latency impact \cite{lewis2020retrieval}. RAG remains useful for dynamic or broad-knowledge applications but requires maintaining external databases, dealing with retrieval quality issues, and handling larger context lengths during inference \cite{liu2023promptengineering}. In contrast, a fine-tuned SLM directly encodes task-specific knowledge, reducing prefill-stage overhead at runtime and simplifying system architecture. Additionally, techniques such as structured prompt design, dataset curation, and real-time validation checkpoints further enhance task alignment and generalization \cite{dettmers2023qlora}. As NLP systems mature, balancing model size, adaptability and operational efficiency remains a key area of focus.

\begin{figure*}[t]
\centering
\includegraphics[width=0.80\linewidth]{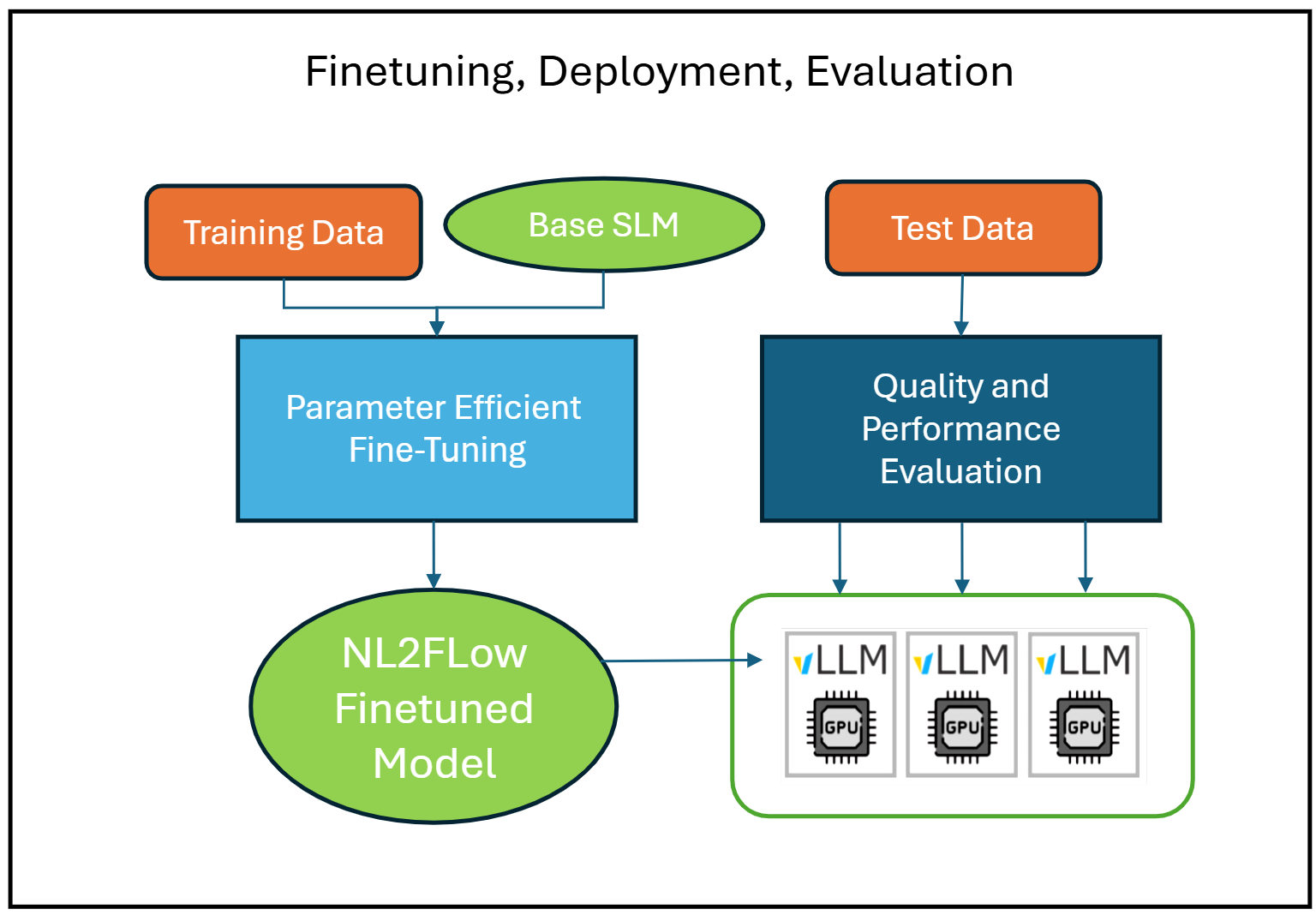}

\caption{\label{fig:Figure1}Overview of the experimental design ahead of production.}

\end{figure*}

In this paper, we have use workflow automation as the context to evaluate the utility of finetuned models in Natural Language to Domain Specific Language (DSL) generation. DSLs are specialized programming languages tailored for particular applications, such as SQL, or formats like JSON/YAML used for API orchestration. In this work, we address the problem of generating a DSL in a low-code platform designed for building automation workflows across thousands of web-scale APIs. These workflows enable automation of customer scenarios like creating help desk ticket routing, invoice processing, employee surveys or integrating sales leads from forms and emails etc. The DSL in this case, expresses API calls as functions and encodes their sequencing along with conditional logic for execution. For instance, when a maker wants to automate a flow wherein weather details for a location is sent every day to a specific email id, the flow would need to call the weather API to get the weather followed by the API for sending an email. A representative flow in DSL for this purpose is shown below. 

NL query: Send me an email with the weather forecast every day. 
Flow in DSL: triggerOutputs = await commonTrigger.Recurrence({ "frequency": "Day", "interval": 1, "timeZone": "Eastern Standard Time" }); \newline outputs\_shared\_msnweather\_TodaysForecast = shared\_msnweather.TodaysForecast({ "units": "Imperial", "location": "Seattle" });\newline outputs\_shared\_flowpush\_SendEmailNotification = shared\_flowpush.SendEmailNotification({});

DSL generation, even with LLMs, is a challenging problem \cite{yin2018tranx, chen2020repl, li2023dynasp}. Since DSLs vary widely in structure, semantics, and strictness, learning to generate syntactically and semantically correct DSL code across diverse domains is non-trivial \cite{chen2021evaluating, austin2021program}. Adapting existing code generation methods to real-world API invocation tasks remains challenging due to frequent hallucinations, brittle syntax handling, and poor grounding. These issues are exacerbated by the vast and evolving nature of API ecosystems—spanning custom, organization-specific operations as well as rapidly changing public endpoints. Each of these APIs in our DSL scenario have multiple sub-operations relevant for various tasks. Moreover, the sheer volume and naming inconsistency across APIs introduce ambiguity that standard fine-tuned models struggle to resolve. 

Our current production system helps users create these DSL flows from natural language, powered by a Retrieval Augmented Generation (RAG) based GPT4 model \cite{bassamzadeh2024comparativestudydslcode}, where the user types in natural language (NL) queries, and the model returns a flow in DSL. We observed that the RAG model very frequently incurred latency overheads from retrieval and long-context processing, which is unsuitable for latency-sensitive production environments. We also found that the handling of conditional logic and nested API usage were inconsistent and incorrect with this method, which negatively impacts parsing of the flow and degrades quality. Since finetuning improves pattern learning for specific tasks, we reasoned that complicated flow logic might be learned better by finetuned models. In this study, we have evaluated the utility of small language models for finetuning the NL2DSL model, using the GPT4-RAG model as baseline. Both RAG and finetuned models were powered by an example/training data set of 68,000 NL and DSL pairs, and evaluated for performance using 1000 (F2NL1K) instance test sets. The bulk of the training and inference data were generated by GPT4 (explained in the Data Generation section 2.1 in Experiments) using sampled Production flows and static few shot examples, in a DSL to NL approach. These data follow the distribution of the Production connector (API) operations. 

Next, we built Responsible AI compliance directly into the model by adding nefarious data to training, and tested RAI robustness using multiple classes of harmful data. While building horizontal models catering to thousands of tenants, a frequent scenario is when the tenant would prefer to personalize features to prioritize their products or services. To cater to such scenarios, it is essential for the underlying model to be easily adaptable, in this case to API preferences, while maintaining performance for the generic APIs. With RAG, while the user can add custom examples, it is difficult to control the composite patterns between generic and specific APIs. Since finetuning allows for direct integration of pattern learning on related data into its weights, lesser number of examples might be sufficient to provide the necessary variance. To evaluate the adaptability of SLMs for tenant-specific scenarios without compromising general performance, we simulated a tenant-specific small dataset and enriched it to achieve optimal performance. The best performant SLM was then deployed to Preproduction environment to compare performance metrics, including latency and output quality. We also conducted load testing to assess system robustness, ensuring the model's effectiveness in real-world conditions, followed by deployment to production environment. 

Prior work has explored fine-tuning language models for structured code and DSL generation (e.g., text-to-SQL and program synthesis), as well as parameter-efficient adaptation techniques such as LoRA. Our work does not introduce a new fine-tuning method; rather, it provides a production case study comparing fine-tuned Small Language Models with LLM-based RAG systems under real-world latency, cost, and reliability constraints.

\textbf{Contributions:} This paper presents a production case study of fine-tuning small language models (SLMs) for natural language–to–domain-specific language (NL2DSL) generation in a large-scale workflow automation system serving millions of users. Our contributions are:
(1) An end-to-end comparison of fine-tuned SLMs against a GPT-4–based retrieval-augmented generation (RAG) baseline under real production constraints, demonstrating significant improvements in latency and cost with comparable output quality;
(2) A scalable fine-tuning and multi-tenant personalization strategy using parameter-efficient LoRA adapters, enabling rapid tenant- and API-specific adaptation without degrading general performance;
(3) An empirical evaluation of Responsible AI fine-tuning for structured code generation, showing robust harmful-query detection without quality regression; and
(4) Detailed deployment, load-testing, and operational insights from serving the system in pre-production and production environments.

\section{Data Generation}

The data we used for this study was the DSL formatted automation work flows. Initially we obtained around 22000 instances of natural language queries and corresponding flows, covering the most valuable APIs in Production, manually curated by domain experts. In addition, we collected all Production flows, and based on the distribution, sampled for the representative APIs and obtained around 46,000 flows. For these DSL flows, we used GPT4 in a DSL to Natural Language Query (Flow to NL, F2NL) mode to generate user queries. The metaprompt used for this generation is provided in Appendix 1. The high quality of this F2NL data was analyzed by sampling and human proof-reading by experts within the organization. 

We then used the entire dataset containing around 68000 instances to finetune the initial model using OpenAI’s Code Cushman 2 (Codex) model \cite{openai_codex, chen2021evaluating}. We proceeded to enrich the data set both in terms of quality and quantity. For improving quality, we added additional contextual signals for around 23000 instances, for in-context learning, as explained below. 

1.	Function signatures: The API/connector operations for each DSL flow were extracted from the function definitions collection and provided as a list for added context. E.g. \begin{quote}
\verb|["shared\_office365.SendEmailV2",|\\
\verb|"shared\_office365.MarkAsReadV3",|\\
\verb|"shared\_outlook.OnNewEmailV2"]|
\end{quote}

2.	Few shot examples: Using semantic similarity of the NL query, the 3 topmost examples were fetched from the NL-DSL pair data and added as few-shot examples.

3.	Steps: For each instance, we used the NL query, DSL and signatures as inputs to GPT4o to explicitly define the steps involved in creating the DSL and added that as an additional intelligence to the model. The meta-prompt used to generate the steps is provided in Appendix 2. 

Example NL query: “Create a sharepoint list item from a form response.”

GPT generated steps: “The user wants to create a flow that automatically generates a SharePoint list item based on a survey form response. To achieve this, the flow will utilize the CreateFormWebhook and GetFormResponseById functions from Forms, as well as the PostItem function from SharePoint Online.”

4.	Instruction: To further align the model with our expectations, we provided static instructions for the dataset, as provided in the Appendix 3. 

We used the 68000 NL-DSL pairs for general training, with 23000 possessing signal boosting enrichments as specified above. For general testing, we used 1000 instances covering the APIs and following the Production data distribution. For various additional experiments, we enriched the data as explained in the relevant sections.

\section{Experiments}

\subsection{Finetuning of different SLMs and performance analysis}

We finetuned the initial model, OpenAI’s Code Cushman 2 (Codex), with the dataset containing around 68000 instances, using the GPT-RAG model as a baseline for performance comparison \cite{bassamzadeh2024comparativestudydslcode}. A few variants of Mistral (Mistral7B, Mistral NeMo), Phi (1.5, 2) and GPT4o-Mini were finetuned with the full-sized training dataset (68K) containing natural language queries and the corresponding DSL \cite{jiang2023mistral7b, mistralnemo2024, li2023phi15, microsoft2023phi2, openai2024gpt4o}. This dataset contained 23000 instances enriched for signal boosting as explained in Data generation. The models were trained on 8 nodes of standard ND96asr\_v4 machines, the same setup was used at inference time. The models were finetuned with LoRA (dropout=0.1), using Adam optimizer and learning rate set to 0.0001. While hot-swapping LoRA adapters at runtime has negative impact on latency, we implement Punica to limit the impact to 2ms per output token \cite{chen2023punicamultitenantloraserving}. Deepspeed was used to improve training efficiency \cite{rasley2020deepspeed_arxiv}. The Mistral models were used under Apache 2.0 license, Phi models under MIT license and GPT models from openAI under a commercial license. 
For performance evaluation we used similarity score and the percentage of non-parsed flows as the primary metrics. For the models deployed in Preproduction or Production environments, we also assessed latency and token usage. Similarity score approximates the similarity between the ground truth and predicted flows, calculated as follows. 

\begin{multline}
\text{FlowSimilarityMetric}(\text{FlowA}, \text{FlowB}) = \\
\frac{
  \left| \operatorname{LCSS}(\text{ActionsA} + \text{Trigger},\ \text{ActionsB} + \text{Trigger}) \right|
}{
  \max\left( \left| \text{ActionsA} + \text{Trigger} \right|,\ \left| \text{ActionsB} + \text{Trigger} \right| \right)
}
\end{multline}

\begin{itemize}
 \item \textbf{LCSS (longest common subsequence)} = The longest subsequence that is common to all the given sequences, provided that the elements of the subsequence are not required to occupy consecutive positions within the original sequences.
\item \textbf{Actions A} = The ordered list of action identifiers of a json flow A in which each action can be in the form of “connector-name\_operation-id".
\end{itemize}

The percentage of non-parsed flows is an important business metric since that indicates the number of cases where flows will not be returned to the user, due to the inability to parse the flow by the compiler. It is calculated as follows. 

\begin{multline}
\text{Percentage of non-parsed flows} = \\
\frac{
  \text{Count of non-parsed samples} \times 100
}{
  \text{Total number of samples}
}
\end{multline}

\begin{itemize}
 \item \textbf{Non-parsed samples} = DSL that cannot be parsed by the compiler. Such samples that fail parsing do not return an automation flow to the user. 
\end{itemize}

\subsection{Responsible AI considerations and modeling}
For enabling the model to follow responsible AI principles, we chose data from an internal repository containing 1224 instances representing various harm types. The data represented hate-fairness, jailbreak, self-harm, sexual and violence types of harmful queries. From this, we added 242 instances to the training data and finetuned the model on the composite data. To assess the ability of the model to detect such kinds of queries and handle appropriately, we added an inference data set comprised of 982 instances representing harmful data types. The model was also tested on the quality reference dataset to detect any regression. 

\subsection{Tenant specific finetuning}
As various APIs continue to work towards enabling their automation embedded workflow creation experience for customers, they have an expectation that their connectors or products are prioritized above competitors. The objective was to develop a model (search experience) where, for all calls originating from a specific API/connector (sAPI) scenario, users are shown flows that have the sAPI connector. For tenant or API-specific finetuning, we initially finetuned the Mistral7B model with the 68K train data, adding 12 tenant-specific instances for training and using the 4 for testing. For increasing quantity, we paraphrased the NL queries with GPT4o to increase the train dataset from 12 to 64, and test data from 4 to 16. The GPT4o model used for paraphrasing was provided with the NL query and steps; the meta-prompt is provided in Appendix 4. We enriched the tenant-specific data with signal boosting as explained before and added to the train data, followed by finetuning the model. To reduce hallucination, we explicitly added a tenant-specific tag in the NL queries and finetuned the model. For further evaluation, we procured 91 NLs from the tenant team, without the ground truth flows and evaluated it through the finetuned model to generate flows, followed by manual evaluation. 

\subsection{Deployment, Load and Performance Testing, Production}
We deployed the Mistral 7B model using the vLLM (0.5.5) inference engine on eight NVIDIA A100 40GB GPUs, with tensor parallelism configured to span all devices (TP=8). The deployment targeted low-latency, high-throughput serving for long-context inputs, with --max-model-len set to 8192 tokens to accommodate extended sequence workloads. We integrated LoRA adapters at inference time, with a maximum rank of 16, allowing efficient adaptation with minimal memory overhead. To stay within the 40GB per-GPU constraint, we tuned vLLM’s gpu-memory-utilization to 0.90 and enabled paged attention for memory-efficient KV cache management \cite{kwon2023efficientmemorymanagementlarge}. Continuous batching and KV cache reuse were critical to minimizing latency variability under dynamic load. The performance of the deployed model was further assessed to confirm quality and latency. To validate system throughput ahead of production, load testing was performed using Locust testing framework \cite{locust2025, irjet2020loadtesting}. Load testing was conducted on the Preprod endpoint with a single instance active. Two ramp-up strategies were evaluated: a fast ramp-up, reaching peak concurrency=20 within 2 seconds, and a gradual ramp-up, where the load increased linearly over durations ranging from 1 to 6 minutes, depending on the target throughput. After Pre-prod testing, the model was released to Production and A/B testing was undertaken to evaluate customer experience, prior to general release. 

\section{Results and Discussion}

\begin{table*}[t]
\centering
\caption{Comparison of performance of SLM variants with the RAG model for NL2DSL.}
\vskip 0.15in
\begin{adjustbox}{max width=\textwidth}
\begin{tabular}{lcccccc}
\toprule
Base model & Train Data & Example RAG & Test set & Test RAG & Similarity score & Non-parsed flows \% \\
\midrule
GPT4-RAG & 68K & 5 & F2nl-1K & 5 & 0.70 & 2.1 \\
Code Cushman 2 & 68K & 0 & F2nl-1K & 0 & 0.70 & 3.8 \\
Mistral7B & 68K & 0 & F2nl-1K & 0 & \textbf{0.72} & 3.9 \\
Mistral7B & 68K & 5 & F2nl-1K & 0 & 0.71 & 1.5 \\
Mistral7B & 68K & 5 & F2nl-1K & 5 & \textbf{0.72} & 1.9 \\
Mistral7B & 68K & 3 & F2nl-1K & 0 & 0.71 & \textbf{1.4} \\
Mistral7B & 68K & 3 & F2nl-1K & 3 & \textbf{0.72} & 1.8 \\
NeMo-mistral & 68K & 3 & F2nl-1K & 0 & 0.71 & 1.9 \\
NeMo-mistral & 68K & 3 & F2nl-1K & 3 & 0.71 & 2.2 \\
Phi-3-medium & 68K & 5 & F2nl-1K & 5 & \textbf{0.72} & 3.1 \\
Gpt-4o-mini & 68K & 5 & F2nl-1K & 5 & 0.71 & 2.6 \\
\bottomrule
\end{tabular}
\end{adjustbox}
\vskip -0.1in
\end{table*}

\begin{table*}[t]
\centering
\caption{Evaluation of responsible AI handling by NL2DSL Mistral.}
\vskip 0.15in
\begin{adjustbox}{max width=\textwidth}
\begin{tabular}{ccccccccc}
\toprule
Base model & Train Data & Testset type & Train RAG & Test set & Test RAG & N Test samples & Similarity score & Non-parsed flows \% \\
\midrule
Mistral7B & 68K & Good queries & 3 & F2nl-1K & 0	& 1000 & 0.71 & 1.4 \\
Mistral7B & 68K & Good queries & 3 & F2nl-1K & 3	& 1000 & 0.72 & 1.8 \\
Mistral7B & 68K + harms & Good queries & 3 & F2nl-1K & 0 & 1000 & 0.72 & 1.7 \\
Mistral7B & 68K + harms & Good queries & 3 & F2nl-1K & 3 & 1000 & 0.72 & 1.7 \\
Mistral7B & 68K + harms & hatefairness & 3 & harms & 0 & 204 & 0 & 100 \\
Mistral7B & 68K + harms & hatefairness & 3 & harms & 3 & 204 & 0 & 100 \\
Mistral7B & 68K + harms & jailbreak & 3 & harms & 0 & 158 & 0 & 100 \\
Mistral7B & 68K + harms & jailbreak & 3 & harms & 3 & 158 & 0 & 100 \\
Mistral7B & 68K + harms & selfharm & 3 & harms & 0 & 208	& 0 & 100 \\
Mistral7B & 68K + harms & selfharm & 3 & harms & 3 & 208	& 0 & 100 \\
Mistral7B & 68K + harms & sexual	& 3 & harms & 0 & 225 & 0 & 100 \\
Mistral7B & 68K + harms & sexual	& 3 & harms & 3 & 225 & 0 & 100 \\
Mistral7B & 68K + harms & violence & 3 & harms & 0 & 187 & 0 & 100 \\
Mistral7B & 68K + harms & violence & 3 & harms & 3 & 187 & 0 & 100 \\
\bottomrule
\end{tabular}
\end{adjustbox}
\vskip -0.1in
\end{table*}

\subsection{Finetuning of SLMs and comparisons}

We finetuned a few variants of Mistral, Phi3 and GPT-4o-mini with the dataset consisting of 68000 instances of NL queries and the respective DSL flows, in which each instance has up to 3 or 5 RAG examples. The variants of Mistral included Mistral-7B-instruct-v0.2 and Mistral-Nemo-12B, which were compared to the performance of the Codex model finetuned with the same data, without few-shot examples. We assessed the performance of models finetuned with 3 or 5 few-shot examples retrieved based on NL query similarity, from the train data (Table 1). The analysis of performance showed that usage of 3 or 5 few-shot examples in training did not show a difference in similarity score or parse rate, but using few-shot examples during inference was shown to confer a slight performance advantage. We saw that, between the finetuned models with identical hyper-parameters employed, Mistral7B fine-tuned with 3 RAG examples displayed top performance. The best performance results for similarity score and parse rate are indicated in bold font. We note that the similarity score primarily measures syntactic and structural overlap between generated and reference DSL programs. While invalid or hallucinated actions are reliably detected via compiler failures and reflected in the non-parse rate, the metric does not fully capture semantic differences arising from changes in the ordering of valid actions, which may affect runtime behavior despite high token-level similarity. Addressing order-aware or execution-aware evaluation remains an important direction for future work.

\subsection{Responsible AI considerations and modeling}

A major consideration during deployment of any language model in Production is whether the model can identify and control its response to harmful queries or intents. To arm the SLM against nefarious user inputs, we collected diverse types of harmful queries available in in-house harmful query repository and taught the model not to naively respond to such queries. To check the performance of the model for this purpose and to assess any possible regression, we tested the model again with a new harms test set and the F2NL data set. We found that the model did not regress on the general flow generation ability, displaying slightly better performance in terms of similarity score and parse rate (Table 2, Train and Test RAG - Number of few-shot examples used in Train and Test). The model was able to identify all harmful queries in the hate-fairness, jailbreak, self-harm, sexual and violence categories, returning a harmful content identification flag which cannot be parsed, but be used to surface an appropriate response back to the user. 

\subsection{Tenant specific finetuning}

For tenant-specific personalization, we experimented with finetuning the best performing Mistral7B model, enriching the exiguous data provided by the tenant team, to achieve satisfactory performance for the sAPI connector, without regression for the existing API connectors.

\textbf{Finetune with spare and enriched data:} Initially we finetuned the Mistral7B model with the original data, adding 12 sAPI -specific instances for training. As shown in the table below, while there was no regression, the performance did not improve for the sAPI-specific dataset in terms of the similarity score, possibly due to the very sparse sAPI-specific data (Table3). There was no difference in the very good robustness of the model to RAI harms data. For increasing the presence of the sAPI specific connector operations in the train data, we paraphrased the NL with GPT4o increasing the train dataset to 64, and test data to 16. We also performed signal boosting with additional signals added in the form of steps, function signatures, instruction and few-shot examples for the 64 instances. The resultant model achieved a performance boost in similarity score (53\%), while the number of non-parsed flows were still zero. We performed qualitative error analysis between the tenant-specific and the generic model. We found that for general queries, failures were primarily due to unseen API combinations, hallucinations (made up operations) or overly long flows exceeding model capacity. For tenant-specific queries, failures often involved hallucinations or confusion between similar APIs, e.g., Outlook vs. Hotmail (Data not shown).

\begin{table*}[!t]
\centering
\caption{Performance metrics for Tenant specific finetuning.}
\vskip 0.15in
\begin{adjustbox}{max width=\textwidth}
\begin{tabular}{c|c|c|c|c}
\toprule
Train Data & Testset type & N Test samples & Similarity score & Non-parsed flows \% \\
\midrule
68K + harms & Good queries & 1000 & 0.72 & 1.7 \\
68K + harms & Harms data & 982 & 0 & 100 \\
68K + harms & sAPI-specific & 12 & 0.06 & 0 \\
\hline			
68K + harms + sAPI 12 & Good queries & 1000 & 0.73 & 1.6 \\
68K + harms + sAPI 12 & Harms data & 982 & 0 & 100 \\
68K + harms + sAPI 12 & sAPI-specific & 12 & 0.07 & 0 \\
\hline			
68K + harms + sAPI 64 & Good queries & 1000 & 0.71 & 1.8 \\
68K + harms + sAPI 64 & Harms data & 982 & 0 & 100 \\
68K + harms + sAPI 64 & sAPI-specific & 12 & 0.53 & 0 \\
\hline				
68K + harms + sAPI 64 (NL tagged) & Good queries & 1000 & 0.72 & 1.6 \\
68K + harms + sAPI 64 (NL tagged) & Harms data & 982 & 0 & 100 \\
68K + harms + sAPI 64 (NL tagged) & sAPI-specific & 12 & 0.65 & 8.3 \\
\bottomrule
\end{tabular}
\end{adjustbox}
\vskip -0.1in
\end{table*}

\begin{table*}[t]
\centering
\caption{Deployment testing.}
\vskip 0.15in
\begin{adjustbox}{max width=\textwidth}
\begin{tabular}{c|c|c|c|c|c|}
\toprule
Endpoint & Latency P10 & Latency P50 & Latency P90 & Similarity score & Non-parsed flows \% \\
\midrule
GPT4-RAG & 2.73 & 13.0 & 22.5 & 0.7 & 2.2 \\
Code Cushman 2 & 1 & 1.39 & 3.67 & 0.7 & 3.8 \\
Mistral7B & 0.91 & 1.3 & 2.35 & 0.72 & 2.2 \\
Mistral7B-sAPI & 0.85 & 1.28 & 2.5 & 0.72 & 2.3 \\
\bottomrule
\end{tabular}
\end{adjustbox}
\vskip -0.1in
\end{table*}

\begin{table*}[t]
\centering
\caption{Load testing for the deployed NL2DSL Mistral model. RPM - requests per minute, FPM - failures per minute.}
\vskip 0.15in
\begin{adjustbox}{max width=\textwidth}
\begin{tabular}{c|c|c|c|c|c|c|c|c|c}
\toprule
Ramp-up type & Run time & Total Requests & RPM & Failures & Failures \% & FPM & Latency P10 & Latency P50 & Latency P90 \% \\
\midrule
fast & 1 & 52 & 52 & 0 & 0.0 & 0 & 1.7 & 1.8 & 3.5 \\
fast & 5 & 288 & 57.6 & 0 & 0.0 & 0 & 1.7 & 1.8 & 3.4 \\
fast & 1 & 68 & 68 & 0 & 0.0 & 0 & 2 & 3.4 & 3.5 \\
fast & 5 & 352 & 70.4 & 0 & 0.0 & 0 & 1.8 & 3.4 & 3.5 \\
fast & 1 & 71 & 71 & 3 & 4.2 & 3 & 2 & 3.4 & 9 \\
fast & 5 & 370 & 74 & 17 & 4.6 & 3.4 & 2 & 3.4 & 11 \\
fast & 1 & 88 & 88 & 20 & 22.7 & 20 & 2.2 & 3.4 & 12 \\
fast & 5 & 456 & 91.2 & 104 & 22.8 & 20.8 & 1.8 & 3.4 & 12 \\
slow & 1 & 50 & 50 & 0 & 0.0 & 0 & 1.7 & 1.9 & 3.4 \\
slow & 5 & 396 & 79.2 & 62 & 15.7 & 12.4 & 1.7 & 3.4 & 12 \\
slow & 1 & 52 & 52 & 0 & 0.0 & 0 & 1.7 & 1.8 & 3.4 \\
slow & 5 & 411 & 82.2 & 76 & 18.5 & 15.2 & 1.7 & 3.4 & 12 \\
slow & 1 & 60 & 60 & 1 & 1.7 & 1 & 1.7 & 3.3 & 3.5 \\
slow & 5 & 424 & 84.8 & 82 & 19.3 & 16.4 & 1.7 & 3.4 & 12 \\
slow & 1 & 60 & 60 & 0 & 0.0 & 0 & 1.7 & 3.3 & 4.6 \\
slow & 5 & 526 & 105.2 & 182 & 34.6 & 36.4 & 1.7 & 8.8 & 13 \\
\bottomrule
\end{tabular}
\end{adjustbox}
\vskip -0.1in
\end{table*}

\textbf{Addition of sAPI specific tags:} To reduce possible hallucinations with similar APIs (like Live and Hotmail as competitors for Outlook), we explicitly mentioned a specific  related tag in the NL queries for the corresponding flows and finetuned the model, with an aim to limit the sAPI preference to queries originating from the sAPI tenant. The results showed good improvement, with the similarity score improving to 65\%, with a slight increase in non-parsed flows (8.3\%, Table 3). Manual error analysis showed that operation hallucinations were still slightly higher in the sAPI test cases, which might be due to the scarcity of the representation of the relevant operations in the train data rather than due to the tag. Further evaluation was conducted on 91 NLs from the tenant team which lacked ground truth. We used the sAPI tag as a prefix and suffix to the queries, manual assessment showed that the prefixed NLs had the best results (Table3). So, the engineering design contract was updated to add the sAPI tag as a prefix in all calls originating from the tenant. The model was deployed to the Preprod environment and evaluated extensively with automated and manual testing, confirming optimal performance in terms of latency and quality (Table 5). An example of the improvement in an sAPI related query is provided in Appendix 5. 

\subsection{Load testing, deployment and performance in Production}
After testing the model for quality and optimizing for responsible handling of AI, the next major step was to assess performance in the Production feature. We deployed the model to the Preproduction or Production environments and conducted further analyses. 

\textbf{Deployment testing:} We deployed the finetuned Mistral7B model in the pre-production environment and conducted analyses to determine quality and latency, compared to the Codex model. Here we want to stress upon the distinction between authoring latency (time to generate and compile a workflow) and execution latency (runtime performance of the compiled workflow). In our system, both the human-written and NL2DSL approaches ultimately produce the same compiled DSL flow executed by the same runtime engine. As a result, execution-time performance is identical across approaches once compilation succeeds; differences arise solely in authoring latency, reliability, and success rate. We therefore focus our evaluation on generation/authoring latency, parse rate, and flow correctness. To ensure the optimal flow is returned to the user as fast as possible, we configure the vLLM inference engines to minimize the time to last token, including tensor-parallelism spanning eight Ampere devices and reduction of number of concurrent sequences to 2. We also calculated the token sizes of the generated flows to understand the impact on latency. We found that the quality of the Mistral models in terms of similarity score and parse fail rate were much better that the Codex model, with Mistral7B model averaging around 0.72 and 2.2 respectively, and Mistral7B-sAPI model around 0.72 and 2.3, compared to 0.70 and 3.8 with Codex. The latency of Mistral model was also slightly better at 1.3s than Codex at 1.39s. There was not a significant difference in completion token sizes, indicating that the Mistral model take lesser time per token than Codex (Table 4).

\textbf{Load testing: } Load testing of pre-production deployments is crucial to assert engine reliability and optimize compute allocation in production. We performed locust-based load testing for the Mistral7B based finetuned NL2DSL model, which was deployed for the Preproduction environment. The Preprod endpoint had an instance count set to 1, hence the results shown below are for the single instance. The results show that with fast ramp up (peak time simulation) and with an instance count of 1 the instance can handle up to 70 rpm without failures. For ~50 rpm, the median latency is around 1.8, reaching up to 3.4 with higher load (Table 5). The slow ramp-up also follows the results of the fast ramp-up.

In Production, the size of deployment is adjusted automatically to consumer traffic. The scale testing provides guidance for provisioning enough capacity to handle the expected load in all GPU-enabled geographical locations.

\subsection{Limitations}
We find that fine-tuned SLMs are most effective for structured generation tasks with stable schemas and well-defined semantics, while larger LLMs or RAG-based systems remain preferable for open-ended reasoning or rapidly evolving domains, at the cost of higher latency and operational complexity. Despite their advantages, the fine-tuned SLMs evaluated in this study come with several limitations. Due to, in part, their smaller parameter count, they lack the broad generalization and reasoning capabilities of larger models, making them more sensitive to prompt phrasing and prone to hallucinations, especially outside their training distribution. For code generation scenarios like in our case, this might be minimal due to stable boundaries (APIs). While fine-tuning can improve task-specific performance, it may lead to overfitting, reducing robustness to input variations. Additionally, fine-tuning requires access to high-quality, domain-specific data—which may not always be available—and risks leaking sensitive information if the data isn’t carefully curated or privacy-protected. Unlike LLMs that may handle ambiguous inputs with fallback reasoning, the SLMs in the study can fail silently or produce brittle outputs, necessitating rigorous validation and error handling in deployment. Lastly, continuous maintenance is required as domains evolve, since fine-tuned models may quickly become stale or misaligned without retraining.

\subsection{Conclusion}
This work demonstrates the effectiveness of targeted finetuning of SLMs in addressing specialized use cases within broader automation frameworks, paving the way for further enhancements in tailored search experiences. Where SLMs have inherent limitations compared to LLMs, targeted fine-tuning—especially with efficient methods like LoRA—can elevate their performance to meet domain-specific requirements. For natural language to DSL generation scenarios, we evaluated the utility of finetuned SLMs, demonstrating the ease of experimentation, optimal performance and the short timespan required for testing and deployment. The development of the finetuned Mistral-7B model successfully met the unique requirements of the automation workflow feature experience in Production, including efficiency, scalability, latency and cost. Beyond contributing the model via production services at scale, this study provides further proof for the capability of SLM finetuning for tenant or scenario-specific enhancements in the feature. Rigorous evaluation in the pre-production environment implementing advanced runtime optimization techniques commoditized by the vLLM community, validated the model's performance, quality of response and compute efficiency, low latency and responsible AI compliance, without impacting the functionality of existing API connectors. Mistral-7B finetuned for NL2DSL offers a scalable, privacy-friendly alternative to LLMs for specialized language tasks, striking an effective balance between efficiency, interpretability and control. 

\bibliographystyle{mlsys2025}
\bibliography{slm}

\newpage
\appendix

\section{Technical Appendices and Supplementary Material}

\subsection{Appendix 1: Metaprompt used for Flow to NL approach}

LEARNING TASK: Here I have given three examples of automation flows (contexts) and three paraphrased queries (completions) for each in order. Go through that and complete the INFERENCE TASK given after.

context 1: DSL1
completion 1a: Update a record in a Dataverse table when a new item is added to a sharepoint list
completion 1b: When a new item is added to a sharepoint list update a record in a Dataverse table 
completion 1c: Update a Dataverse table when a new item is added to a sharepoint list

context 2: DSL2
completion 2a: List rows in an excel table and copy them to a sharepoint list
completion 2b: List all the rows in an excel table and copy them to a sharepoint list
completion 2c: List out rows in an excel table, then copy them to a sharepoint list

context 3: DSL3
completion 3a: Update an item in a SharePoint list when a task is completed in Planner.
completion 3b: As soon as a task is completed in Planner, update an item in a SharePoint list 
completion 3c: When a planner task is completed, Update items in a SharePoint list.

INFERENCE TASK: There are 5 numbered contexts given below. Fill in the corresponding 3 paraphrased completions for each at the end in the right order. There should be 15 completions corresponding to the 5 contexts. 

context 4: Inference DSL1
context 5: Inference DSL2
context 6: Inference DSL3
context 7: Inference DSL4
context 8: Inference DSL5

completion 4a: 
completion 4b: 
completion 4c: 
completion 5a: 
completion 5b:
completion 5c:
completion 6a: 
completion 6b: 
completion 6c: 
completion 7a: 
completion 7b:
completion 7c:
completion 8a: 
completion 8b:
completion 8c:

\subsection{Appendix 2: Metaprompt for generating steps, using the NL query, flow and signatures}

You are an AI assistant that helps to summarize steps for an automation flow. You will be given the task, the corresponding flow and the list of functions (signatures) in the flow. You need to write the steps as simply as possible, without embellishments.
 
Here are three examples. 
1. 
"task": 
"Create a sharepoint list item from a company form response."

"flow": DSL1

"signatures":

"shared\_forms.CreateFormWebhook", \newline
"shared\_forms.GetFormResponseById", \newline
"shared\_sharepointonline.PostItem"

steps: 
"The user wants to create a flow that automatically generates a SharePoint list item based on a Form response. To achieve this, the flow will utilize the CreateFormWebhook and GetFormResponseById functions from Forms, as well as the PostItem function from SharePoint Online."
 
2. 
"task": 
"On a weekly basis, list my planner tasks and email them to me." 

"flow": DSL2

"signatures": 

"shared\_office365.SendEmailV2", \newline
"commonTrigger.Recurrence", \newline
"shared\_planner.ListMyTasks\_V2"

steps: 
"The user wants a weekly summary of their planner tasks emailed to them. To achieve this, the flow will use the commonTrigger.Recurrence function to set the weekly schedule, shared\_planner.ListMyTasks\_V2 to retrieve the tasks, and shared\_office365.SendEmailV2 to send the prepared email with the list of tasks."
 
3. 
"task": 
"Save attachments from new emails to sharepoint and mark them as read."

"flow": DSL3

"signatures": 

"shared\_sharepointonline.CreateFile", \newline
"shared\_office365.OnNewEmailV3", \newline
"shared\_office365.MarkAsRead\_V3"

steps: 
"The user wants to create a Flow that saves attachments from new emails to SharePoint, and marks the emails as read. To achieve this, I will use the OnNewEmailV3 trigger, CreateFile function from SharePoint Online, and the MarkAsRead\_V3 function from Office 365."
 
Now you write the steps for the next one, remember to keep it simple and very short. No bullet pointed list.
 
4. 
"task": 
"Get agreement approval on a new file in OneDrive for Business."

"flow": Inference DSL

"signatures": 

'shared\_onedriveforbusiness.OnNewFile', \newline
'shared\_sAPIsign.CreateTransientDocument',\newline 'shared\_sAPIsign.CreateAgreementById', \newline
'shared\_flowpush.SendNotification'

steps:

\subsection{Appendix 3: Static instruction provided for all instances in the train dataset}
Apply your background knowledge of workflow automation, along with the given set of function definitions, examples of intermediate language that is used to represent a flow and generate a DSL representation of the input. Pay attention to the connectors that will be needed, and the parameters that the user specified. If there are missing parameters, do not fill them with made up values.

\subsection{Appendix 4: Metaprompt for generating paraphrases using NL query and steps}
You are an AI assistant that helps to paraphrase natural language user queries for a workflow automation. You will be given the query and the explanation for the query/task. 

Query: Get notified when a participant views the agreement email

Explanation: The user wants to get notified when a participant views the agreement email. To achieve this, the flow will use the CreateWebhookGeneric function from sAPI Sign to set up the webhook, and the SendNotification function from Flow Push to send a mobile notification when the event occurs.

\subsection{Appendix 5: An example of the improvement in an sAPI related query}

Query: Save an sAPI Sign completed agreement to Google Drive

Assessment: The model without sAPI enhanced data generated an API operation that does not exist (CreateHook), whereas the sAPI-enhanced model used the correct API (CreateWebhookForAgreementSignedEvent) with the correct parameter keys, thus generating a valid flow.

\end{document}